\documentclass[11pt]{article}
\usepackage[a4paper,margin=2.5cm]{geometry}
\usepackage{cite}
\usepackage{url}
\usepackage{authblk}

\begin{document}

\author[*]{David Alvarez L.} 
\author[**]{Monica Iglesias M.}
\renewcommand\Authands{ and }
\title{\emph{k}-Means Clustering and Ensemble of Regressions: An Algorithm for the ISIC 2017 Skin Lesion Segmentation Challenge}
\affil[*]{Medical Physicist at Asturias University Central Hospital, Oviedo, Spain}
\affil[**]{Medical Imaging Technologist}
\date{\today}
\maketitle

\let\thefootnote\relax\footnote{Corresponding author: David Alvarez (david.alvarezl@sespa.es)}

\section{Motivation and overview}
This abstract briefly describes a segmentation algorithm developed for the ISIC \cite{isic} 2017 Skin Lesion Detection Competition hosted at \cite{kitware}. The objective of the competition is to perform a segmentation (in the form of a binary mask image) of skin lesions in dermoscopic images as close as possible to a segmentation performed by trained clinicians, which is taken as ground truth. This project only takes part in the segmentation phase of the challenge. The other phases of the competition (feature extraction and lesion identification) are not considered.

The proposed algorithm consists of 4 steps: (1) lesion image preprocessing, (2) image segmentation using \emph{k}-means clustering of pixel colors, (3) calculation of a set of features describing the properties of each segmented region, and (4) calculation of a final score for each region, representing the likelihood of corresponding to a suitable lesion segmentation. The scores in step (4) are obtained by averaging the results of 2 different regression models using the scores of each region as input. Before using the algorithm these regression models must be trained using the training set of images and ground truth masks provided by the Competition. Steps 2 to 4 are repeated with an increasing number of clusters (and therefore the image is segmented into more regions) until there is no further improvement of the calculated scores.

\section{Description of the segmentation algorithm}
In the following sections each step of the algorithm is described in detail, as well as the workflow between them and the processing of the final segmentation.

\subsection{Image preprocessing}
Prior to segmentation, all images are preprocessed in order to minimize undesirable features that could affect the performance of the algorithm, such as bright reflections, presence of hair and color differences between images. The images are also normalized to a unique size and shape. This normalization allows for the comparison of region features such as positions and sizes between different images.
	
\subsubsection*{Size normalization} 
All images are rescaled to $1024 \times 1024$ pixels using linear interpolation. In order to prevent distortion of the image, the smallest dimension of the image (either height or width) is expanded with zero-valued rows or columns of pixels on both sides before rescaling, so the image becomes square shaped.

\subsubsection*{Removal of reflections}
Reflections in dermoscopic images typically appear as small, very bright areas. In this project the \emph{brightness} of a color pixel is defined as the sum of the values of its 3 channels (red, green and blue). The value of the brightest pixels is replaced by the average of their neighbours, excluding any adjacent pixel also classified as \emph{bright}. In order to decide whether a pixel must be replaced, the following rule is followed: the value $t$ of the 99-percentile of brightness in each image is found; then, all pixels with a brightness greater than $0.98t$ are classified as \emph{bright} and replaced. This threshold was chosen by visually evaluating the results on a small number of images of the training set. 

\subsubsection*{Removal of hair}
The method proposed by \cite{silveira2009} is used. First, a morphological closing \cite{gonzalez2008} operation is performed independently on each of the color channels of the image, using as selection element a disk of radius equal to 5 pixels. Then a median $3\times3$ filter is applied to every channel.

\subsubsection*{White balance}
A simple white balance is performed, based on the \emph{Gray World Theory} \cite{zapryanov2012} assumption. The values of all pixels on each color channel are added. Let us denote the 3 resulting values $S_{red}$, $S_{green}$ and ${S_{blue}}$. Using these values, 2 color balance factors are calculated as $k_{red}=S_{red}/S_{green}$ and $k_{blue}=S_{blue}/S_{green}$. Finally, multiplying the red and blue values of each pixel in the image by their corresponding factor results in an image with an equal ``amount of color'' in all 3 channels.

\subsection{Image segmentation}
The preprocessed image is segmented using an iterative color clustering algorithm, starting with 3 clusters. All pixels in the image (considered as points in a 3-dimensional space whose components are the 3 color channels of the image) are assigned to one of the clusters using a \emph{k}-means algorithm.

For each cluster a binary mask is constructed, showing the locations of the pixels belonging to it. Since the mask can contain many small non-connected regions, morphological opening and closing operations are performed,  using a disk of radius 10 pixels. These operations allow small regions to be absorbed into their bigger neighbours. The number of small, probably not significant regions, is thereby reduced.

\subsection{Calculation of region features}
A set of 10 features is calculated for each of the remaining regions in all clusters. Each feature encodes a different property of the region as a number in the range $[0,1]$. The following features are defined:

\subsubsection*{Region area}
A histogram of 500 bins in the interval $[0, 1024^2]$ is built using the areas of the  lesion masks provided in the training set, previously rescaled to $1024\times1024$. The histogram is normalized to a maximum equal to 1. The value of the histogram for a given region area is then used as the area feature value for that region. Thus, regions are assigned a bigger value of this feature the more frequent their area is in the training set.

\subsubsection*{Region position}
To calculate this feature, the centroids of all rescaled lesion masks in the training set are calculated. A 2-dimensional gaussian is then fit to the set of centroids. The maximum of the gaussian, which is situated very close to the center of the images, is set to 1. The value of the position feature for any region is obtained as the value of the gaussian at the coordinates of the centroid of the region. Thus, regions whose centroid is close to the center of the image are assigned the highest value of this feature, while the value diminishes the further the centroid is from that position.

\subsubsection*{Region circularity}
Let $A$ and $p$ be the area and perimeter of a region respectively. The \emph{circularity} of the region is defined as 
$$circularity=\frac{4\pi A}{p^2}$$
Note that for a perfectly circular region $circularity=1$.

\subsubsection*{Region solidity}
\emph{Solidity} is defined as the ratio of the number of pixels in the region to the number of pixels in the convex hull of the region

\subsubsection*{Region average color (3 channels)}
The mean value of each color channel for every lesion in the training set is calculated. Then the mean and the standard deviation of the values obtained is calculated for each channel. A 1-dimensional gaussian function is defined for each color using the corresponding mean and deviation, with the maximum of the function set to 1. For any given region, the color feature for each channel is calculated by applying these functions to the mean color of that region. Note that those regions with a mean color more similar to the mean color of all lesions in the training set will have a higher value of these features.

\subsubsection*{Similarity between region color and color in the center of the image (3
 channels)}
 The image is divided into 9 equal sized squares, and only the pixels in the central square are considered. The mean and standard deviation of the pixel values in the central area are calculated for each of the 3 color channels separately. Using the same approach as in the previous feature, 3 feature values are calculated which will account for the similarity between the mean color of the region and the mean color in the central area of the image. 
 
\subsection{Calculation of region score}
Using the set of 10 features, a score is calculated for each region applying 2 different regression models. The average of the 2 scores is used as the final score for the region. These scores are predictions of the Jaccard Index between the region and the ground truth mask of the image. The models must be previously trained with the training set provided by the Competition (see section \ref{training}). The models used were:

\begin{itemize}
\item Random forest \cite{breiman2001} composed of 50 random trees
\item Epsilon-Support Vector Regression \cite{smola2004} (SVR) with the following parameters: $C=100$, $\gamma=0.5$, and $\epsilon=0.2$
\end{itemize}

The values for the parameters of both models were chosen by evaluating the performance of different values over a local validation set. This set was formed by a subset of 200 images of the training set. Other regression models such as linear or \emph{k} nearest neighbours were considered, but the combination of random forest and SVR yielded the best results over the validation set.
    
\subsection{Prediction workflow and postprocessing of the final result}
When all regions have been scored, the algorithm stores the highest score found and the corresponding region as the best candidate for the lesion segmentation. A new clustering is performed increasing the number of clusters by 1, and the resulting regions are scored again. This process is repeated until there is no improvement in the scores after increasing the number of clusters. At this point, the iteration stops and the region with the best score is considered the best estimation of the lesion segmentation.

When the best candidate region (herein \emph{mask}) has been determined, the following postprocessing steps are executed:
\begin{enumerate}
\item All ``holes'' (background regions completely surrounded by the mask) are deleted and their pixels are incorporated to the mask.
\item A morphological closing  of the mask is performed using a disk of radius 30 pixels, followed by a dilation \cite{gonzalez2008} using a disk of radius 14. These operations compensate for the fact that this algorithm tends to underestimate the margin to be left around the lesions compared to the ground truth masks. They also smooth the border of the mask. The optimal radii for the disks were found by optimizing over the local validation set.
\end{enumerate}

\section{Training of regression models} \label{training}
The regression algorithms need to be trained before being used for prediction. In order to generate a large and diverse sample set for the training step, the same clustering and feature calculation algorithm described in section 2 was used on the images of the training set, but the scoring step was replaced by a \emph{naive} score calculation; each region was assigned a score equal to the sum of its 10 feature values. For each region generated, its array of feature values were added to the sample set, as well as the Jaccard index of the region compared to the ground truth mask for that image. The set of all pairs of feature values and their associated Jaccard indexes can then be used to train the regression algorithms.

\bibliographystyle{ieeetr}
\bibliography{biblio}

\end{document}